\documentclass[runningheads]{llncs}
\usepackage[T1]{fontenc}
\usepackage{graphicx}
\usepackage{booktabs}
\usepackage[misc]{ifsym}
\usepackage{algorithm}
\usepackage{algorithmicx}
\usepackage{algpseudocode}
\usepackage{amsmath}
\usepackage{amsfonts}
\usepackage{multirow}
\usepackage{subfigure}
\usepackage{threeparttable}
\usepackage{xcolor}
\definecolor{lightblue}{RGB}{100, 160, 255}  % 浅蓝色
\definecolor{wineRed}{RGB}{140, 20, 40}  % 酒红色
\usepackage{enumitem}
\usepackage{diagbox}
\usepackage{float}
\usepackage{dsfont}
\newcommand{\corr}{(\Letter)}
\usepackage{mwe}
\begin{document}

\title{Cross-Domain Conditional Diffusion Models for Time Series Imputation}

%\titlerunning{Underwater Basket Weaving Under Extreme Pressure}
% If the full title of your paper is short enough to also fit in the running head, you can omit the abbreviated paper title here. You can check as follows: if you comment out the \titlerunning line, something will appear in the header of all odd-numbered pages of your PDF from page 3 onward. This something is either the full title (in which case all is well), or the error message "Title Suppressed Due to Excessive Length". If this error message appears, you're going to want to provide an abbreviated title within the \titlerunning command, because if you won't do it, Springer will do it for you.

%N.B.: Author information (both in the \author{} and \authorrunning{} command) should only be present in the Camera-Ready Version of your paper. The version that you initially submit for review, ought to be double-blind. So, when initially submitting your paper, use:
%\author{Author information scrubbed for double-blind reviewing}
\author{Kexin Zhang\inst{1}\thanks{Kexin Zhang is a research intern during the completion of this work.} \and
Baoyu Jing\inst{2} \and
K. Selçuk Candan\inst{3} \and
Dawei Zhou\inst{4} \and
Qingsong Wen\inst{5} \and
Han Liu\inst{1} \and
Kaize Ding\inst{1} \corr}
% You may leave out the orcidID information, if you want to.
% Use \corr to indicate the corresponding author. Note the spacing around the \corr command. Only one author can be the corresponding author.

%N.B.: comment out the \authorrunning{} command for the double-blind version of your paper submitted for review. Later, if your paper is accepted, use the command for the Camera-Ready Version.
\authorrunning{K. Zhang et al.}
% If there are two authors, write 'A.L. Benjamin and C.C. Broadus Jr.'
% If there are more than two authors, '[...] et al.' is used.

\institute{Northwestern University, Evanston, IL 60208, USA \email{kevin.kxzhang@gmail.com, \{hanliu, kaize.ding\}@northwestern.edu}
\and
University of Illinois Urbana-Champaign, Champaign, IL 61820-5711, USA \email{baoyuj2@illinois.edu}
\and
Arizona State University, Tempe, AZ 85281, USA \email{candan@asu.edu}
\and
Virginia Tech, Blacksburg, VA 24061-0131, USA \email{zhoud@vt.edu}
\and
Squirrel Ai Learning, Bellevue, WA 98004, USA \email{qingsongedu@gmail.com}
}
\maketitle              % typeset the header of the contribution

\begin{abstract}
Cross-domain time series imputation is an underexplored data-centric research task that presents significant challenges,
particularly when the target domain suffers from high missing rates and domain shifts in temporal dynamics.
%Time series imputation faces critical challenges in cross-domain scenarios, particularly when addressing high missing rates in the target domain and 
%discrepancies in temporal dynamics across domains.
%distribution shifts between source and target data.
%A common approach is to mix all available data and train a single model across domains. However, this can lead to suboptimal imputation, as domain-specific patterns are not properly captured, and the model struggles to adapt to the target domain.
Existing time series imputation approaches primarily focus on the single-domain setting, which cannot effectively adapt to a new domain with domain shifts.
%While diffusion models have been explored for time series imputation due to their strong generative capabilities, existing methods often construct imputation targets by treating missing values as zeros. This disrupts the continuity of temporal patterns and prevents the model from leveraging cross-domain information.
%, making it difficult to align imputed values with the target domain distribution. 
%Furthermore, conventional consistency regularization methods enforce alignment between source and target domains by directly matching predictions, without considering domain-specific variations and the uncertainty introduced by missing data, which can lead to suboptimal adaptation and degraded imputation quality when the imputed values are unreliable.
Meanwhile, conventional domain adaptation techniques struggle with data incompleteness, as they typically assume the data from both source and target domains are fully observed to enable adaptation. For the problem of cross-domain time series imputation, missing values introduce high uncertainty that hinders distribution alignment, making existing adaptation strategies ineffective.
%Furthermore, conventional domain adaptation techniques struggle with incomplete data, as they typically assume fully observed distributions for alignment. In time series imputation, missing values introduce uncertainty that distorts distribution alignment, making rigid alignment strategies ineffective.
%Conventional domain adaptation techniques enforce alignment through direct prediction matching, which overlook domain-specific variations and fail to account for uncertainty introduced by missing data.
%Additionally, missing data introduces uncertainty, which conventional consistency regularization fails to address, as it enforces alignment through direct prediction matching without considering domain-specific variations. This can lead to error accumulation and unreliable imputations in the target domain.
% To address these challenges, we propose 
% %\kz{why CD2-TSI? give a hint}
% a \textbf{C}ross-\textbf{D}omain \textbf{C}onditional \textbf{D}iffusion model for  \textbf{T}ime \textbf{S}ries \textbf{I}mputation (\textbf{CD$^2$-TSI}). 
%\kx{should we follow the order of data, model, algo here?}
Specifically, our proposed solution tackles this problem from three perspectives: 
\textbf{(i) Data:} We introduce a frequency-based time series interpolation strategy that integrates shared spectral components from both domains while retaining domain-specific temporal structures, constructing informative priors for imputation. 
\textbf{(ii) Model:} We design a diffusion-based imputation model that effectively learns domain-shared representations and captures domain-specific temporal dependencies with dedicated denoising networks.
\textbf{(iii) Algorithm:} We further propose a cross-domain consistency alignment strategy that selectively regularizes output-level domain discrepancies, enabling effective knowledge transfer while preserving domain-specific characteristics.
Extensive experiments on three real-world datasets demonstrate the superiority of our proposed approach. Our code implementation is available here\footnote{\url{https://github.com/kexin-kxzhang/CD2-TSI}}.

\end{abstract}

\section{Introduction}
\label{sec:intro}
%1. Background
Multivariate time series imputation is essential for various real-world applications, including environmental monitoring and energy management~\cite{zhou2021informer}.
Missing values commonly arise due to sensor failures, transmission errors, or external disruptions, leading to incomplete data that could degrade the reliability of downstream tasks~\cite{little2019statistical}.
Effective imputation is thus critical for preserving the integrity of the data and ensuring reliable results in subsequent applications~\cite{kong2025time}.

Many endeavors have been made to model the temporal patterns inherent in time series. 
Traditional statistical and machine learning methods for time series imputation often assume stationarity or linear relationships, which may not capture the full complexity of real-world time series data. 
Recurrent neural networks and attention-based models have improved the modeling of temporal dependencies by capturing nonlinear relationships~\cite{cao2018brits}.
More recently, deep generative models~\cite{yoon2018gain,fortuin2020gp}, such as variational autoencoders and generative adversarial networks, have been explored for time series imputation. 
Diffusion-based models~\cite{tashiro2021csdi,alcaraz2022diffusion,liu2023pristi,zhou2024mtsci} further advance imputation by learning a denoising process that iteratively refines missing values.

Despite their success, these methods struggle under high missing rates~\cite{tashiro2021csdi}, as sparse observations hinder the effective modeling of the underlying temporal dependencies~\cite{du2024tsi,zhou2024mtsci,gao2024diffimp}.
When observations are highly incomplete, it is natural to leverage related domains to improve imputation performance~\cite{ding2021cross}. For instance, in air quality monitoring, neighboring cities' sensor networks may provide complementary temporal patterns when local sensors fail.
Recent advancements in domain adaptation (DA) have shown promising results in transferring knowledge across domains in tasks such as time series forecasting and classification~\cite{jin2022domain,ragab2023adatime}. In light of this, we propose to tackle the novel cross-domain time series imputation by adapting domain discrepancies between two related domains.

However, cross-domain time series imputation remains largely underexplored and directly applying DA techniques to time series imputation may easily fail due to the following challenges:
\textbf{(1) \textit{Data Challenge:}} %Domain shifts exacerbated by data incompleteness.
Most of the DA problems usually assume the observed data from both source and target domains are complete~\cite{zhang2024survey}, however, in time series imputation, it is hard for existing models to well characterize the real data distributions due to the high missingness in the observed data. 
\textbf{(2) \textit{Model Challenge}:} 
Existing approaches commonly rely on simply training a single shared model on mixed data, which cannot distinguish domain-shared and specific knowledge~\cite{zhang2024incmsr}. Therefore, it is necessary to develop a model that can both facilitate knowledge transfer and capture domain-specific patterns.
\textbf{(3) \textit{Algorithm Challenge}:} 
Time series data from different domains often exhibit domain-specific temporal dependencies, leading to variations in seasonality, trend shifts, or periodic patterns~\cite{yue2022ts2vec}, etc. Existing domain adaptation algorithms often enforce alignment without considering such differences across domains, thus failing to capture the cross-domain knowledge required for accurate imputation in the target domain.

To address these challenges, we propose a novel \textbf{C}ross-\textbf{D}omain \textbf{C}onditional \textbf{D}iffusion Model for \textbf{T}ime \textbf{S}eries \textbf{I}mputation \textbf{(CD$^2$-TSI)}, which improves imputation in the target domain by leveraging knowledge from a source domain while preserving domain-specific temporal patterns.
Specifically, to counter the data challenge, we introduce a frequency-based time series interpolation strategy, which interpolates original missing values by integrating shared spectral components from both the source and target domains. The pre-interpolated values are used to construct the 
missing targets, 
providing more informative priors for training the imputation model.
For the model challenge, we develop a diffusion-based imputation framework that learns domain-shared representations to capture common patterns across domains while maintaining dedicated denoising networks to model domain-specific temporal dependencies.
To tackle the algorithm challenge, we propose a cross-domain consistency alignment algorithm that imposes alignment based on output-level discrepancy. The degree of alignment is adjusted according to the prediction difference between source and target networks for the same target samples. 
This approach facilitates cross-domain transfer while preserving target-specific temporal characteristics, preventing the model from overfitting to source domain patterns.

In summary, the main contributions of this work are summarized as follows: (1) We target the problem of cross-domain time series imputation, which is largely underexplored and requires research attention in the community. 
(2) We propose CD$^2$-TSI, a new diffusion model-based framework that solves the problem of cross-domain time series imputation from data, model, and algorithm perspectives.
(3) We conduct extensive experiments on three real-world datasets, demonstrating that CD$^2$-TSI outperforms state-of-the-art models across various missing data patterns, highlighting its effectiveness in cross-domain settings.

\section{Related Work}
\label{sec:related_work}

\subsection{Time Series Imputation}
Time series imputation (TSI) methods can be broadly categorized into predictive and generative approaches~\cite{du2024tsi}:
(1) Predictive methods~\cite{wu2022timesnet,du2023saits,marisca2022learning} predict deterministic values but suffer from error accumulation and fail to capture the uncertainty of missing values. GRU-D~\cite{che2018recurrent} and BRITS~\cite{cao2018brits} use deep autoregressive models with time decay, while GRIN~\cite{cini2021filling} incorporates graph neural networks (GNN) for spatial relationships.
(2) Generative methods, such as those based on Variational Autoencoders (VAE), Generative Adversarial Networks (GAN), and diffusion models, effectively circumvent the limitations faced by those predictive models. VAE-based methods~\cite{miyaguchi2022variational,kim2023probabilistic} optimize reconstruction error and regularize the latent space. 
GAN-based approaches~\cite{miao2021generative} use adversarial training between the generator and discriminator but can be unstable and produce unrealistic results. Diffusion models show promise due to their ability to model complex data distributions and generate varied outputs for missing values. CSDI~\cite{tashiro2021csdi} and SSSD~\cite{alcaraz2022diffusion} use observed data as conditional information;  PriSTI~\cite{liu2023pristi} extracts conditional information and considers spatiotemporal dependencies using geographic data;
MTSCI~\cite{zhou2024mtsci} incorporates a complementary mask strategy and a mixup mechanism to realize intra-consistency and inter-consistency.
However, these methods focus primarily on modeling temporal dependencies within a single domain, overlooking the complexities posed by cross-domain scenarios where domain shifts in missing patterns or temporal dynamics exist. Our CD$^2$-TSI framework addresses this gap by combining diffusion models with domain adaptation to enhance imputation quality across domains.

\subsection{Time Series Domain Adaptation}
Domain Adaptation (DA)~\cite{eldele2024label} seeks to transfer knowledge to a target domain by leveraging information from source domains. 
These methods can be categorized into three groups:
(1) Adversarial-based methods train a domain discriminator to identify domains while learning transferable features. For example, CoDATS~\cite{wilson2020multi} employs a gradient reversal layer for adversarial training with weak supervision on multi-source data. SLARDA~\cite{ragab2022self} aligns temporal dynamics across domains via autoregressive adversarial training.
(2) Discrepancy-based methods use statistical distances to align features from source and target domains. AdvSKM~\cite{liu2021adversarial} leverages maximum mean discrepancy (MMD) with a hybrid spectral kernel for temporal domain adaptation. RAINCOAT~\cite{he2023domain} tackles feature and label shifts via temporal and frequency feature alignment.
(3) Self-supervision methods incorporate auxiliary tasks. DAF~\cite{jin2022domain} uses a shared attention module for domain-invariant and specific features and reconstruction.
While existing DA methods have proven effective in forecasting and classification tasks, their application to imputation remains underexplored, where temporal discrepancies as well as data deficiency introduced by missing values pose additional challenges. 
CD$^2$-TSI differs from these approaches by addressing the challenges introduced by incomplete observations in cross-domain time series imputation and adaptation.

\section{Problem Definition}

Cross-domain time series imputation aims to reconstruct missing values in a target domain by leveraging knowledge from a related source domain.
Given multivariate time series data with potential missing values, we define the time series in both domains as
$\mathbf{X}_{d} = (\mathbf{X}_{d,1}, \ldots, \mathbf{X}_{d,K}) \in \mathbb{R}^{K \times L}$, where $K$ is the number of features, $L$ is the length of the time series, and $d \in \{Src, Tgt\}$ denotes the source and target domains. We assume all time series in both domains have the same length. 
An observation mask $\mathbf{M}_{d} \in \{0, 1\}^{K \times L}$ indicates missing values, where $m_{k,l} = 1$ if the value is observed for the $k$-th feature at the $l$-th timestamp, and $m_{k,l} = 0$ if the value is missing.
Since real-world datasets often lack ground truth for missing data, we artificially mask a subset of observed values for training and evaluation. Following previous work~\cite{zhou2024mtsci,tashiro2021csdi}, the extended missing targets $\widetilde{\mathbf{X}}_{d} \in \mathbb{R}^{K \times L}$ include both the original missing values and artificially masked values, with a binary mask $\widetilde{\mathbf{M}}_{d} \in \{0, 1\}^{K \times L}$.

\section{Methodology}
\label{sec:method}
\subsection{Model Overview (CD$^2$-TSI)}
\begin{figure*}[ht]
\centering
\includegraphics[width=0.99\linewidth]{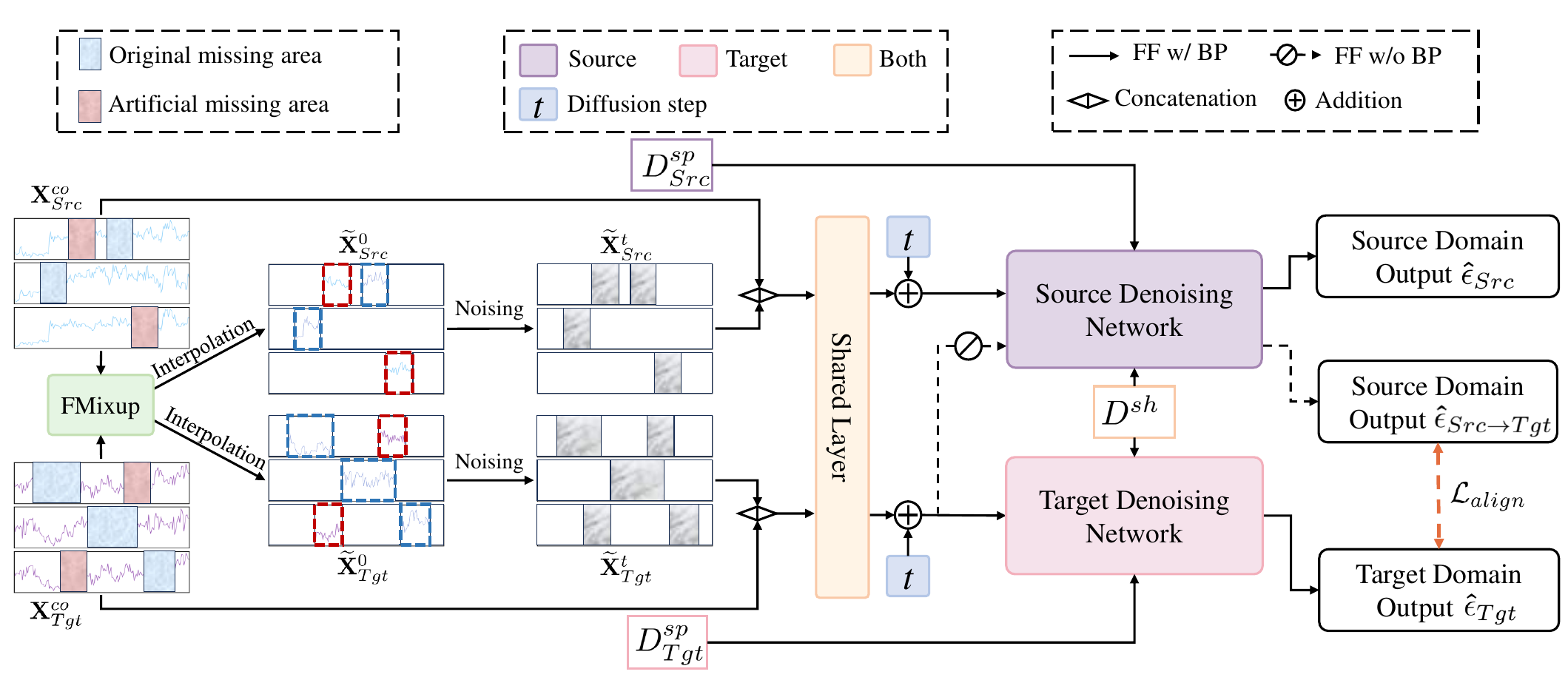} 
\caption{Architecture of CD$^2$-TSI. FMixup is utilized to interpolate the original missing areas (\textcolor{lightblue}{blue}), while artificial missing values (\textcolor{wineRed}{red}) are retained to construct the missing targets $\widetilde{\mathbf{X}}^0$. These targets are then transformed into the noisy targets $\widetilde{\mathbf{X}}^t$ to train the denoising network, with the help of conditional information $\mathbf{X}^{co}$. The framework is optimized using a combination of denoising loss and consistency alignment loss.}
\label{fig:framework}
\end{figure*}

As shown in Fig.~\ref{fig:framework}, CD$^2$-TSI incorporates a cross-domain diffusion-based framework, where source and target domains share representations while maintaining domain-specific denoising networks.
A pre-interpolation strategy is proposed to integrate spectral components from both domains, providing priors for original missing values via cross-domain frequency mixup, while 
the artificial missing values
are retained to construct 
missing targets.
These targets are then corrupted by adding noise to obtain noisy inputs
for training the denoising network.
To ensure effective adaptation,
we introduce cross-domain consistency alignment.
This algorithm promotes adaptation based on output discrepancy while preventing excessive regularization that could force the target domain to overly conform to source domain patterns.
Overall, CD$^2$-TSI effectively leverages cross-domain information to improve the imputation quality in the target domain. 

\subsection{Conditional Diffusion Model for Time Series Imputation}
Imputing missing values in time series data requires capturing complex temporal dependencies while addressing challenges from data 
incompleteness.
Our framework takes Denoising Diffusion Probabilistic Models (DDPM)~\cite{ho2020denoising} as the base model, where the imputation process is formulated as a conditional generative task. 
In the forward process, Gaussian noise is step by step added to the 
missing targets
$\widetilde{\mathbf{X}}^0$ across $T$ diffusion steps, gradually transforming $\widetilde{\mathbf{X}}^0$ into a noisy version $\widetilde{\mathbf{X}}^T$. 
This process is formalized as follows:
\begin{equation}
q\left(\widetilde{\mathbf{X}}^{1: T} \mid \widetilde{\mathbf{X}}^0\right)=\prod_{t=1}^T q\left(\widetilde{\mathbf{X}}^t \mid \widetilde{\mathbf{X}}^{t-1}\right), q\left(\widetilde{\mathbf{X}}^t \mid \widetilde{\mathbf{X}}^{t-1}\right)=\mathcal{N}\left(\widetilde{\mathbf{X}}^t ; \sqrt{1-\beta_t} \widetilde{\mathbf{X}}^{t-1}, \beta_t \boldsymbol{I}\right)
\end{equation}
where $\beta_t$ 
represents the noise level,
and $t$ indicates the diffusion step. According to DDPM, 
$\widetilde{\mathbf{X}}^t=\sqrt{\bar{\alpha}_t} \widetilde{\mathbf{X}}^0+\sqrt{1-\bar{\alpha}_t} \epsilon$, 
where $\alpha_t=1-\beta_t, \bar{\alpha}_t=\prod_{i=1}^t \alpha_i$, and $\epsilon \sim \mathcal{N}(\mathbf{0}, \mathbf{I})$ where $\mathcal{N}$ is Gaussian distribution. When $T$ is large enough, $q\left(\widetilde{\mathbf{X}}^{T} \mid \widetilde{\mathbf{X}}^0\right)$ approximates a standard normal distribution.

The reverse process then reconstructs the 
missing targets
by iteratively denoising imputed values, conditioned on the remaining observations $\mathbf{X}^{co}$:
\begin{equation}
p_\theta\left(\widetilde{\mathbf{X}}^{t-1} \mid \widetilde{\mathbf{X}}^t, \mathbf{X}^{co} \right)=\mathcal{N}\left(\mu_\theta\left(\widetilde{\mathbf{X}}^t, \mathbf{X}^{co}, t\right), \sigma_t^2 \boldsymbol{I}\right),
\end{equation}
\begin{equation}
\mu_\theta\left(\widetilde{\mathbf{X}}^t, \mathbf{X}^{co}, t\right)=\frac{1}{\sqrt{\bar{\alpha}_t}}\left(\widetilde{\mathbf{X}}^t-\frac{\beta_t}{\sqrt{1-\bar{\alpha}_t}} \epsilon_\theta\left(\widetilde{\mathbf{X}}^t, \mathbf{X}^{co}, t\right)\right), \sigma_t^2=\frac{1-\bar{\alpha}_{t-1}}{1-\bar{\alpha}_t} \beta_t
\end{equation}
where $\epsilon_\theta(\cdot)$ is the denoising network with learnable parameters $\theta$.
The model is trained to estimate the added noise $\epsilon$ given $\widetilde{\mathbf{X}}^t$, conditional observations $\mathbf{X}^{co}$ and current diffusion step $t$, and  
the training objective of time series imputation is:
\begin{equation}
\mathcal{L}(\theta)=\mathbb{E}_{\widetilde{\mathbf{X}}^0 \sim q\left(\widetilde{\mathbf{X}}^0\right), \epsilon \sim \mathcal{N}(0, I)}\left\|\epsilon-\epsilon_\theta\left(\widetilde{\mathbf{X}}^t,  \mathbf{X}^{co}, t\right)\right\|^2
\label{eq:loss}
\end{equation}

\subsection{Cross-Domain Time Series Frequency Interpolation}
Modeling temporal dependencies from incomplete time series is challenging, especially under high missing rates.
Severe missingness disrupts the real data distribution, making it difficult for the model to capture consistent temporal dependencies.
Addressing this issue is crucial, as many existing methods~\cite{tashiro2021csdi,alcaraz2022diffusion} simply replace original missing values with \textit{zeros} when constructing the 
missing targets.
However, \textit{zeros} cannot reflect the real data distribution, and such a distribution shift makes it more challenging for the diffusion model to accurately recover missing values during the denoising process.
Although linear interpolation in the time domain can partially address this issue, it often fails to capture complex non-linear temporal dynamics. 

To solve this problem, our intuition is that 
time series data from related domains typically share low-frequency components, which represent long-term trends or periodic patterns (e.g., daily cycles in hydrology data), while high-frequency components reflect domain-specific details, such as sensor noise or transient fluctuations~\cite{xu2021fourier,zhang2022self,yang2024frequency}. 
Formally, a signal can be decomposed into an amplitude spectrum, which captures the intensity of different frequency components, and a phase spectrum, which preserves local temporal structure. 
Hence, we propose a frequency-based time series interpolation strategy -- FMixup.
FMixup is achieved through two key steps: (1) blending \textit{low-frequency amplitude} spectra across domains and (2) retaining each domain’s \textit{high-frequency amplitude} and \textit{phase} spectra. The augmented data can then be used to replace original missing values and refine the missing targets.

\smallskip
\noindent \textbf{Domain-Shared Frequency Mixup.} To exchange structural information across domains, we transform the conditional observations $\mathbf{X}^{co} \in \mathbb{R}^{K \times L}$ from both 
domains into the frequency domain using the Fast Fourier Transform (FFT)~\cite{nussbaumer1982fast}:
\begin{equation}
\mathcal{F}\left(x\right)(u, v)=\sum_{k=0}^{K-1} \sum_{l=0}^{L-1} x(k, l) e^{-j 2 \pi\left(\frac{k}{K} u+\frac{l}{L} v\right)}
\end{equation}
where $u$ and $v$ are frequency indices along the two dimensions, and $j$ is the imaginary unit.
This frequency space signal $\mathcal{F}\left(x\right)$ can be further decomposed into an amplitude spectrum $\mathcal{A} \in \mathbb{R}^{K \times L}$ and a phase spectrum $\mathcal{P} \in \mathbb{R}^{K \times L}$. 
To integrate
common patterns,
we introduce a binary mask $ \mathcal{M}=\mathds{1}_{(k, l) \in [-\alpha K : \alpha K, -\alpha L : \alpha L]} $ that selects low-frequency region of the amplitude spectrum, where $\alpha \in (0,1)$ determines the proportion of low-frequency information incorporated. 
The amplitude spectra of the source and target domains are then blended as follows:
\begin{equation}
\mathcal{A}_{Src \rightarrow Tgt}=\mathcal{A}_{Tgt} * (1-\mathcal{M}) + (\lambda  \mathcal{A}_{Tgt} + (1-\lambda) \mathcal{A}_{Src}) * \mathcal{M}
\end{equation}
where $\mathcal{A}_{Src}$ and $\mathcal{A}_{Tgt}$ represent the amplitude spectra of the source and target domains, respectively. $\mathcal{A}_{Src \rightarrow Tgt}$ is the newly mixed amplitude spectrum, and parameter $\lambda$ adjusts the balance between the two spectra.

\smallskip
\noindent \textbf{Domain-Specific Frequency Preserving.}
As mentioned above, high-frequency components often contain domain-specific fine-grained details. To preserve such information, we retain the high-frequency amplitude components of the target domain.
Additionally, we do not modify the phase spectrum $\mathcal{P}_{Tgt}$, as it represents local structural information necessary for preserving the original sequence characteristics.
The final 
%reconstructed 
augmented
time series is obtained by combining the mixed amplitude spectrum $\mathcal{A}_{Src \rightarrow Tgt}$ with original phase spectrum $\mathcal{P}_{Tgt}$ of the target domain and applying the inverse Fourier transform:
\begin{equation}
X_{Src \rightarrow Tgt}=\mathcal{F}^{-1}\left(\mathcal{A}_{Src \rightarrow Tgt}, \mathcal{P}_{Tgt}\right)
\end{equation}

This augmented series is used to fill in the original missing values in the target domain.
Therefore, we obtain the refined 
%imputation targets 
missing targets
$\widetilde{\mathbf{X}}^0_{Tgt}$. 
This frequency-based time series interpolation strategy ensures that the local temporal structure remains aligned with the target domain while benefiting from shared low-frequency trends.
Similarly, we obtain the 
missing targets
$\widetilde{\mathbf{X}}^0_{Src}$.

\subsection{Cross-Domain Conditional Diffusion Model}
Although cross-domain time series frequency mixup provides priors from the data perspective, it does not fully address 
domain shifts in 
temporal dynamics.
Learning a single shared model for both source and target domains often fails to capture domain-specific patterns, leading to suboptimal imputation performance in the target domain. 
To address this, we propose a novel cross-domain conditional diffusion model that enables domain-shared knowledge transfer while modeling domain-specific temporal dependencies.

\smallskip
\noindent \textbf{Domain-Shared Temporal Knowledge Transfer.}
To facilitate knowledge transfer across domains, we try to learn domain-shared input representations using a shared convolution layer. The input representations for the source and target domains are formulated as:
$H^{in}_{Src}=\operatorname{Conv}\left(\mathbf{X}^{co}_{Src} \| \widetilde{\mathbf{X}}^t_{Src}\right)$ and $H^{in}_{Tgt}=\operatorname{Conv}\left(\mathbf{X}^{co}_{Tgt} \| \widetilde{\mathbf{X}}^t_{Tgt}\right)$, respectively, where $\operatorname{Conv}$ is $1 \times 1$ convolution. 
To further integrate shared information and help the imputation, the model incorporates domain-shared side information $D^{sh}$, which includes: (1) a time embedding $s = \{s_1, \dots, s_L\} \in \mathbb{R}^{L \times 128} $ for temporal dependencies, constructed using sine-cosine temporal encoding~\cite{vaswani2017attention}; (2) a learnable feature embedding $f = \{f_1, \dots, f_K\} \in \mathbb{R}^{K \times 16}$ to model shared feature relationships. We expand and concatenate $s$ and $f$ and obtain $D^{sh} \in \mathbb{R}^{K \times L \times C}$, where $C$ is the channel size.

\smallskip
\noindent \textbf{Domain-Specific Temporal Knowledge Modeling.}
After the common feature extraction, the model applies domain-specific attention mechanisms:
\begin{equation}
\begin{aligned}
& H^{tem}=\operatorname{Attn}_{tem}\left(H^{in} + \operatorname{Linear}(t_{emb})\right) \\
& H^{fea}=\operatorname{Attn}_{fea}\left(H^{tem}\right)
\end{aligned}
\end{equation}
where $\operatorname{Attn}_{tem}(\cdot)$ captures temporal dependencies, and $\operatorname{Attn}_{fea}(\cdot)$ models feature interactions. These attention layers are domain-specific, enabling the model to learn unique characteristics within each domain. The diffusion step embedding $t_{emb}$
is constructed through sine-cosine temporal encoding as well and projected through a linear layer.
Additionally, domain-specific side information $D^{sp}$ includes the conditional mask $\widetilde{\mathbf{M}}$ of each domain, which explicitly indicates missing positions.
The final output of each denoising network is computed as:
\begin{equation}
H^{out} = H^{fea} + \operatorname{Conv}(D^{sh}) + \operatorname{Conv}(D^{sp}).
\end{equation}

The domain-specific modeling stacks multiple layers, where the output $H^{out}$ of each layer is divided into a residual connection and a skip connection after a gated activation unit. The residual connection serves as the input to the next layer, while the skip connections from each layer are summed and passed through two layers of $1 \times 1$ convolution to obtain the final output.
\subsection{Cross-Domain Consistency Alignment (CDCA)}
To mitigate temporal discrepancies across domains while accounting for the uncertainty caused by missing values, we further propose cross-domain consistency alignment.
Unlike conventional domain adaptation methods,
which enforce rigid alignment regardless of the magnitude of domain discrepancies, CDCA 
selectively enforces prediction consistency based on the model output discrepancy for the same target domain samples.

Let $\mathbf{\hat{\epsilon}}_{Tgt}$ denote the target network's prediction on a given target sample, and let $\mathbf{\hat{\epsilon}}_{Src \rightarrow Tgt}$ denote the prediction from the source network (in evaluation mode) when the same target sample is used as input.
The average absolute difference between these predictions, denoted as $\Delta$, is then computed: 
\begin{equation}
\Delta=\frac{1}{N} \sum_{i=1}^N\left\|\mathbf{\hat{\epsilon}}_{Tgt, i}-\mathbf{\hat{\epsilon}}_{Src \rightarrow Tgt, i}\right\|
\end{equation}
where $N$ is the number of target domain samples.

CDCA compares $\Delta$ against two thresholds: a lower threshold $\tau_l$ and an upper threshold $\tau_h$. If $\Delta<\tau_l$, the discrepancy is within an acceptable range, where enforcing alignment could amplify the impact of missingness-induced noise rather than improving adaptation. If $\tau_l \leq \Delta \leq \tau_h$, the discrepancy is moderate. In this case, we impose a penalty proportional to the excess difference, specifically $\Delta-\tau_l$. If $\Delta>\tau_h$, the discrepancy is large, suggesting that strict alignment could cause overfitting to source domain patterns and distort intrinsic target structures. To prevent this, the penalty is capped at $\min(\Delta-\tau_l, \tau_h)$.
Thus, the alignment loss is formulated as:
\begin{equation}
\mathcal{L}_{align}= \begin{cases}0, & \Delta<\tau_l, \\ \min \left(\Delta-\tau_l, \tau_h\right), & \Delta \geq \tau_l .\end{cases}
\label{eq:L_ca}
\end{equation}

By applying regularization only when discrepancies exceed a lower threshold and capping penalties for large deviations,
CDCA achieves a trade-off between imposing cross-domain alignment and preserving target-specific characteristics.

\subsection{Overall Loss Function}

The overall loss function for our model integrates several components: time series imputation losses (Eq.~\ref{eq:loss}) for both source and target domains, and an auxiliary loss that addresses cross-domain consistency alignment (Eq.~\ref{eq:L_ca}), re-weighted by parameters $\mu_{align}$. The overall loss function is defined as:
\begin{equation}
\mathcal{L}=\mathcal{L}_{Src}+\mathcal{L}_{Tgt}+\mu_{align}\mathcal{L}_{align}
\label{eq:L_total}
\end{equation}

This formulation ensures that the model not only learns to impute missing values within each domain but also mitigates domain discrepancies.
\section{Experiments}
\label{sec:experiment}

\subsection{Experimental Setting}

We evaluate our model on three real-world datasets, with details described below and the statistics of datasets are presented in Table~\ref{tab:datasets}.

\textbf{Air Quality}~\cite{yi2016st} dataset contains PM2.5 measurements from Beijing (B) and Tianjin (T). Beijing data is collected from 36 stations, while Tianjin has data from 27 stations. For cross-domain setting, 27 stations with the fewest missing values were sampled from Beijing data.

\textbf{Hydrology} dataset records daily river flow and sediment concentration from 20 stations in the United States, collected from United States Geological Survey~\cite{Konrad2022} and Water Quality Portal~\cite{read2017water}. It consists of two domains: Discharge (D) and Pooled (P), spanning from March 1, 2017, to September 30, 2022.

\textbf{Electricity}~\cite{zhou2021informer} dataset consists of power load and oil temperature data.
It includes two years of data (from July 2016 to July 2018) from two distinct regions in China, referred to as ETTh1 and ETTh2.

We follow the dataset splitting strategy used in prior work~\cite{zhou2024mtsci,tashiro2021csdi,liu2023pristi}. For Air Quality dataset, we select Mar., Jun., Sep., and Dec. as the test set, the last 10\% of the data in Feb., May, Aug., and Nov. as the validation set, and the remaining data as the training set. For Hydrology and Electricity, we split the training/validation/test set by 70\%/10\%/20\%.

\textbf{Evaluation Metrics.}
We evaluate the performance using three metrics: Mean Absolute Error (MAE), Root Mean Squared Error (RMSE), and Continuous Ranked Probability Score (CRPS). MAE and RMSE measure the error between imputed values and ground truth for deterministic methods. CRPS is used to measure how well the imputed probability distributions align with the observed values for methods that produce probability distributions.

\textbf{Masking Strategy.}
Since original missing values within datasets lack ground truth, we consider two missing patterns to simulate the missing values for evaluation: (1) \textit{Point missing}, where 10$\%$ of the observations is masked, and (2) \textit{Block missing}, where we mask 5 $\%$ of the observed data and mask observations ranging from 1 to 4 data points for each feature with 0.15 $\%$ probability. For training strategies, we use two masking strategies for self-supervised learning:  (1) \textit{Point strategy} randomly selects $r (r \in [0\%, 100\%])$ of observed values; (2) \textit{Block strategy} randomly select a sequence of length $[L/2, L]$ as 
missing targets
with an additional 5 $\%$ of observed values randomly selected.
Since Air Quality dataset has much original missing data in the training set, we adopt point missing pattern following previous work~\cite{tashiro2021csdi}. For Hydrology and Electricity, we apply both point and block missing patterns following~\cite{liu2023pristi}.
\begin{table}[!t]
\caption{Dataset Characteristics}
\centering
\label{tab:datasets}
\scalebox{0.85}{ 
\setlength{\tabcolsep}{1.2pt} 
\renewcommand{\arraystretch}{1.2} 
\fontsize{9}{11}\selectfont 
\begin{tabular}{ccccccc}
\toprule
\multirow{2.5}{*}{\diagbox[width=2.5cm]{Statistics}{Dataset}} & \multicolumn{2}{c}{Air Quality} & \multicolumn{2}{c}{Hydrology} & \multicolumn{2}{c}{Electricity} \\
\cmidrule(lr){2-3} \cmidrule(lr){4-5} \cmidrule(lr){6-7}
 & Beijing & Tianjin & Discharge & Pooled & ETTh1 & ETTh2 \\
\midrule
Samples & 8759 & 8759 & 2726 & 2726 & 17420 & 17420 \\
Length & 36 & 36 & 16 & 16 & 48 & 48 \\
Features & 27 & 27 & 20 & 20 & 7 & 7 \\
Original Missing Rate & 12.36\% & 20.84\% & 0\% & 19.99\% & 0\% & 0\% \\
\bottomrule
\end{tabular}
} 
\end{table}

\textbf{Baselines and Implementation Details.}
\textbf{Baselines for DA.} The chosen baselines include various state-of-the-art methods that have been widely adopted in the time series classification and forecasting tasks: CORAL~\cite{sun2016deep}, CDAN~\cite{long2018conditional}, DIRT-T~\cite{shu2018dirt}, AdvSKM~\cite{liu2021adversarial}, CotMix~\cite{eldele2023contrastive}.
\textbf{Baselines for TSI.} The baselines include RNN-based models M-RNN~\cite{yoon2018estimating}, BRITS~\cite{cao2018brits}, GNN-based models GRIN~\cite{cini2021filling}, SPIN and SPIN-H~\cite{marisca2022learning} and diffusion-based methods CSDI~\cite{tashiro2021csdi}, SSSD~\cite{alcaraz2022diffusion}, PriSTI~\cite{liu2023pristi}, and MTSCI~\cite{zhou2024mtsci}.
\textbf{Hyperparameters.} We set the batch size to 16 and the number of epochs to 200. The Adam optimizer is used with an initial learning rate of 1e-3, decaying to 1e-4 and 1e-5 at 75\% and 90\% of the total epochs, respectively. The frequency space mix ratio $\lambda$ is sampled within [0.0, 1.0], and $\alpha$ in FMixup is empirically set as 0.003. As for the model, we use 4 residual layers, 64 residual channels ($C$), and 8 attention heads. For methods requiring an adjacency matrix, we use the identity matrix by default. 
We adopt the quadratic schedule for other noise levels following~\cite{tashiro2021csdi}, with a minimum noise level $\beta_1 = 0.0001$ and a maximum noise level $\beta_T = 0.5$.

\subsection{Overall Performance}
The overall comparisons on three datasets are shown in Table~\ref{tab:overall}. We summarize the observations as follows: 
(1) CD$^2$-TSI consistently outperforms baseline models across all datasets. 
Compared with existing time series imputation methods that rely solely on single-domain data, 
CD$^2$-TSI integrates cross-domain knowledge, leading to better imputation performance.
Additionally, CD$^2$-TSI outperforms domain adaptation methods by specifically handling discrepancies in temporal dynamics and data deficiency caused by missing values, which are not adequately addressed by conventional DA approaches.
(2) Incorporating domain adaptation techniques increases imputation accuracy compared to training solely on the target domain. 
However, the extent of this improvement varies across different missing patterns. For example, CDAN performs well on point missing but falls short on block missing scenarios due to its alignment strategy.
In contrast, CD$^2$-TSI consistently improves upon the strongest DA baselines, 
with an average improvement of 1.92\% (RMSE) and 1.34\% (MAE), demonstrating its effectiveness in cross-domain alignment for imputation.
(3)  Our method achieves notable improvements over the best TSI baselines. Specifically, on the Air Quality dataset, CD$^2$-TSI provides a +4.41\% improvement in RMSE and a +5.46\% improvement in MAE. On the Hydrology dataset, it results in a +4.37\% improvement in RMSE and a +4.61\% improvement in MAE, while on the Electricity dataset, they are +4.13\% in RMSE and +2.04\% in MAE. Among TSI models, MTSCI's performance varies across datasets, with lower accuracy on Air Quality and Hydrology due to high missing rates.
These results underscore CD$^2$-TSI's capacity to adapt effectively to different real-world datasets even under severe missing conditions.

\begin{table*}[!t]
\caption{The overall performance comparison. Bold scores are the best performance, and underlined scores are the best time series imputation baseline performance.}
\scalebox{0.85}{
\centering
\setlength{\tabcolsep}{4pt} 
\renewcommand{\arraystretch}{1.2} 
\footnotesize 
\begin{tabular}{ccccccccccc}
\toprule
\multirow{4}{*}{Method} & \multicolumn{2}{c}{Air Quality} & \multicolumn{4}{c}{Hydrology} & \multicolumn{4}{c}{Electricity } \\
& \multicolumn{2}{c}{\textit{B→T}} & \multicolumn{4}{c}{\textit{D→P}} & \multicolumn{4}{c}{\textit{h1→h2}} \\
& \multicolumn{2}{c}{Point} & \multicolumn{2}{c}{Point} & \multicolumn{2}{c}{Block} & \multicolumn{2}{c}{Point} & \multicolumn{2}{c}{Block} \\
\cmidrule(lr){2-3} \cmidrule(lr){4-5} \cmidrule(lr){6-7} \cmidrule(lr){8-9} \cmidrule(lr){10-11}
 & RMSE & MAE & RMSE & MAE & RMSE & MAE & RMSE & MAE & RMSE & MAE\\
\midrule
Coral    & 14.814 & 7.374 & 48.353 & 13.579 & 45.484 & 16.370 & 0.645 & 0.381 & 1.230 & 0.548 \\
CDAN     & 14.594 & 7.203 & 47.968 & 13.370 &  44.725 & 16.403 & 0.644 & 0.381 & 1.270 & 0.563 \\
Dirt-T   & 14.945 & 7.429 & 48.506 & 13.659 &  45.394 & 16.227 & 0.672 & 0.396 & 1.194 & 0.559 \\
AdvSKM   & 14.786 & 7.311 & 48.776 & 13.289 &  43.996 & 16.012 & 0.643 & 0.380 & 1.283 & 0.567 \\
CoTMix   & 14.632 & 7.232 & 48.470 & 13.685 &  44.810 & 15.954 & 0.651 & 0.383 & 1.235 & 0.568 \\ \midrule
M-RNN    & 46.226 & 29.497 & 58.975 & 19.465 &  47.000 & 21.524 & 7.338 & 5.386 & 11.309 & 4.428 \\
BRITS    & 40.067 & 26.355 & 56.749 & 19.249 &  47.207 & 22.251 & 6.988 & 4.428 & 8.109 & 4.893 \\
GRIN     & 26.274 & 15.773 & 60.845 & 23.690 &  55.254 & 27.094 & 3.744 & 1.587 & 3.273 & 1.854\\
SPIN     & 27.881 & 16.914 & 59.263 & 22.020 &  53.530 & 25.428 & 6.750 & 2.856 & 7.096 & 3.503 \\
SPIN-H   & 30.895 & 18.617 & 58.915 & 19.501 &  49.234 & 19.893 & 6.947 & 2.941 & 8.001 & 6.064 \\
CSDI     & \underline{15.002} & \underline{7.452} & \underline{48.542} & \underline{13.744} &  \underline{45.819} & \underline{16.470} & \underline{0.647} & \underline{0.380} & 1.320 & 0.592 \\
SSSD     & 15.536 & 8.086 & 49.132 & 14.668 &  47.079 & 18.282 & 0.787 & 0.501 & 1.307 & 0.677 \\
PriSTI   & 15.546 & 7.686 & 48.760 & 14.064 &  47.868 & 18.032 & 0.683 & 0.407 & 1.329 & 0.615 \\
MTSCI   & 16.252 & 8.793 & 48.941 & 14.477 & 48.094 & 18.523 & 0.715 & 0.483 & \underline{1.240} & \underline{0.562} \\
\textbf{CD$^2$-TSI} & \textbf{14.339} & \textbf{7.045} & \textbf{46.852} & \textbf{13.182}&  \textbf{43.407} & \textbf{15.626} & \textbf{0.635} & \textbf{0.378} & \textbf{1.161} & \textbf{0.542} \\
\bottomrule
\end{tabular}
}
\label{tab:overall}
\end{table*}
\subsection{Sensitivity and Ablation Study}
\noindent \textbf{Sensitivity Analysis.}
We investigate the sensitivity of our method to different missing rates and masking strategies. We conduct experiments on the Air Quality dataset to evaluate performance under various missing rates and on the Hydrology and Electricity datasets to assess different masking strategies.

\begin{figure*}[ht]
\centering
\includegraphics[width=0.85\linewidth]{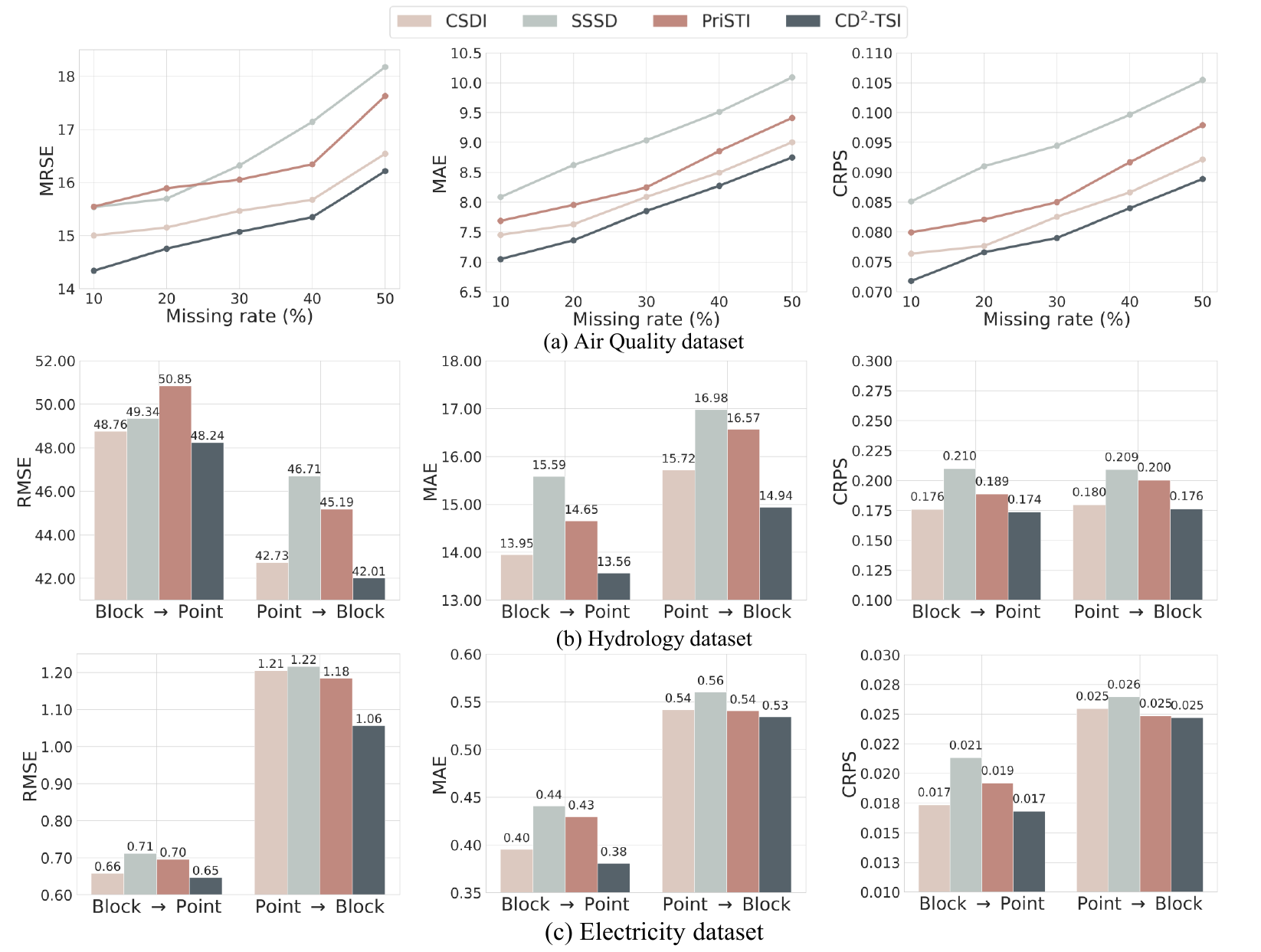} 
\caption{Sensitivity Analysis for Air Quality, Hydrology, Electricity Datasets}
\label{fig:sensitivity}
\end{figure*}

For the Air Quality dataset, we randomly select 10/20/30/40/50\% of the observed values as ground truth in the test data. Fig.~\ref{fig:sensitivity} (a) shows that our method consistently performs well across these rates. As missing data increases, imputation accuracy typically declines due to reduced availability of observed conditional information. 
However, our approach's use of frequency mixup interpolation and cross-domain alignment helps maintain high performance.

For the Hydrology and Electricity datasets, we use two settings: Point $\rightarrow$ Block (Point missing pattern in training set, Block missing pattern in testing set) and Block $\rightarrow$ Point (Block missing pattern in training set, Point missing pattern in testing set). Fig.~\ref{fig:sensitivity} (b-c) shows that CD$^2$-TSI achieves relatively better performance with various missing patterns in the training and testing sets.

\noindent \textbf{Ablation Study.}
We evaluate the impact of key components in CD$^2$-TSI on imputation performance.
\textbf{(1) w/o FMixup}: FMixup is excluded, and zero filling is used to construct the 
missing targets.
\textbf{(2)} \textbf{w/ L.I.}: FMixup is replaced with linear interpolation to construct the 
missing targets.
\textbf{(3) w/o CDCA}: Consistency alignment loss $L_{align}$ is removed.
Since Electricity dataset does not contain original missing values, no interpolation is required for original missing areas, and we only evaluate w/o CDCA on this dataset.
\begin{table*}[!t]
\caption{Ablation Study of CD$^2$-TSI.}
\scalebox{0.83}{
\centering
\setlength{\tabcolsep}{4pt} 
\renewcommand{\arraystretch}{1.2} 
\footnotesize 
\begin{tabular}{ccccccccccc}
\toprule
\multirow{4}{*}{Method} & \multicolumn{2}{c}{Air Quality} & \multicolumn{4}{c}{Hydrology} & \multicolumn{4}{c}{Electricity} \\
& \multicolumn{2}{c}{\textit{B→T}} & \multicolumn{4}{c}{\textit{D→P}} & \multicolumn{4}{c}{\textit{h1→h2}} \\
& \multicolumn{2}{c}{Point} & \multicolumn{2}{c}{Point} & \multicolumn{2}{c}{Block} & \multicolumn{2}{c}{Point} & \multicolumn{2}{c}{Block} \\
\cmidrule(lr){2-3} \cmidrule(lr){4-5} \cmidrule(lr){6-7} \cmidrule(lr){8-9} \cmidrule(lr){10-11}
 & RMSE & MAE & RMSE & MAE & RMSE & MAE & RMSE & MAE & RMSE & MAE \\
\midrule
w/o FMixup & 14.817 & 7.292 & 47.662 & 13.365 & 44.204 & 15.956 & - & - &  - & - \\
w/ L.I. & 14.921 & 7.357 & 48.524 & 13.557 & 44.261 & 16.163 &  - & - &  - & - \\
w/o CDCA & 14.782 & 7.319 & 47.288 & 13.390 & 43.679 & 15.876 & 0.641 & 0.380 & 1.255 & 0.554 \\
\textbf{CD$^2$-TSI} & \textbf{14.339} & \textbf{7.045} & \textbf{46.852} & \textbf{13.182} & \textbf{43.407} & \textbf{15.626}& \textbf{0.635} & \textbf{0.378} & \textbf{1.161} & \textbf{0.542}\\
\bottomrule
\end{tabular}}
\label{tab:ablation}
\end{table*}

Table~\ref{tab:ablation} presents the results of our ablation study. 
Removing frequency mixup interpolation (w/o FMixup) significantly degrades performance across all datasets and missing patterns, confirming that frequency mixup provides informative priors that enhance imputation accuracy. When FMixup is replaced with linear interpolation (w/ L.I.), performance further declines, demonstrating that frequency-domain interpolation captures temporal dependencies more effectively than simple interpolation in the time domain.
Removing the cross-domain consistency alignment loss (w/o CDCA) results in performance degradation, particularly in block missing scenarios. For instance, in the Electricity dataset, RMSE increases from 1.161 to 1.255 in block missing pattern. This confirms that cross-domain consistency alignment helps in 
mitigating temporal discrepancies across domains.
Overall, the findings of the ablation studies underscore the importance of each proposed component in improving cross-domain imputation.

\subsection{Hyperparameter and Efficiency Study}
\begin{figure*}[ht]
\centering
\includegraphics[width=0.85\linewidth]{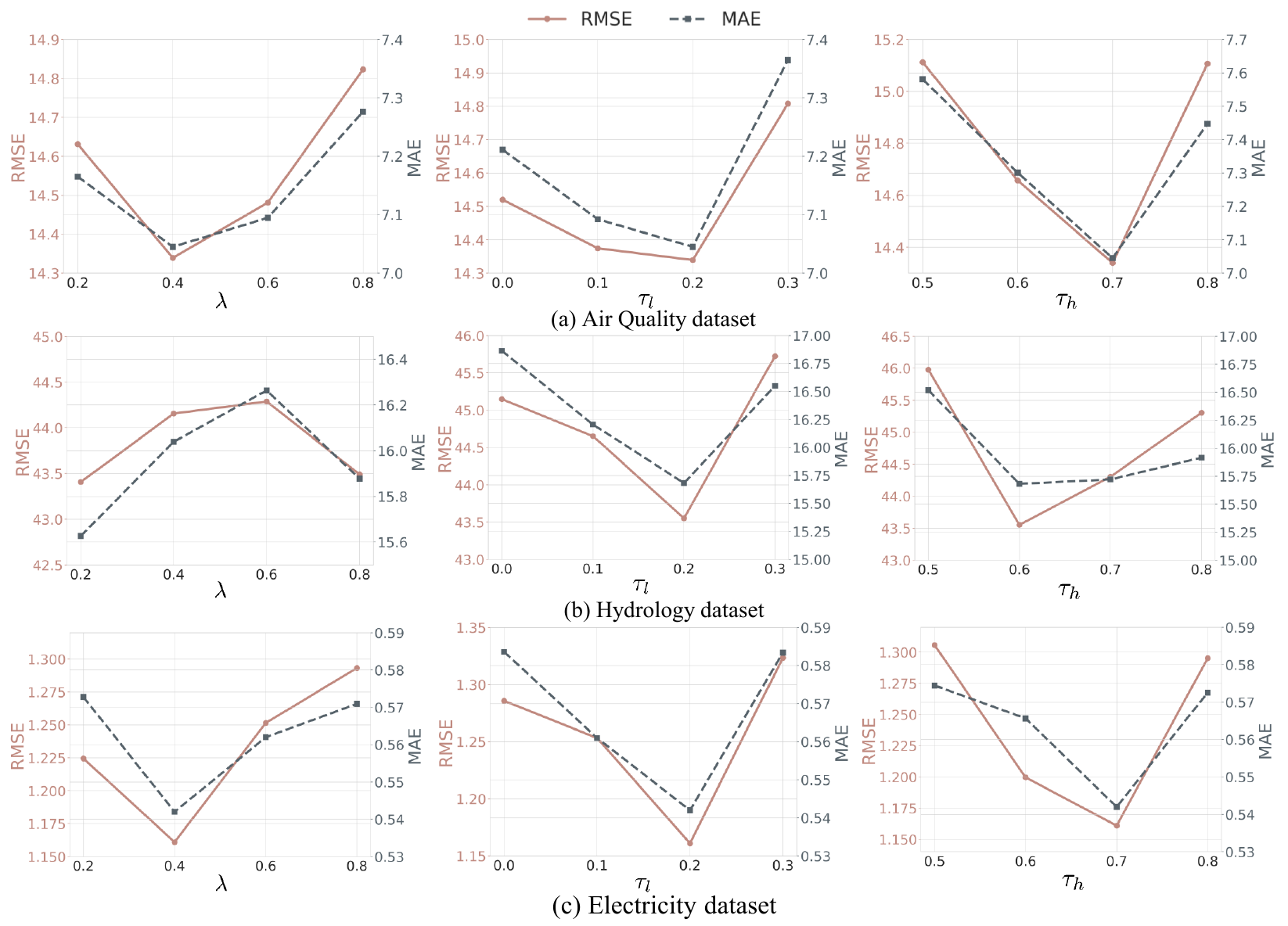} 
\caption{Hyperparameter study on three key parameters of CD$^2$-TSI.}
\label{fig:hyper}
\end{figure*}

\noindent \textbf{Hyperparameter Study.}
We conduct a hyperparameter study on key parameters in CD$^2$-TSI to select the optimal settings across three datasets: the frequency space mix ratio $\lambda$, the lower and upper thresholds $\tau_l$ and $\tau_h$. The results are shown in Fig.~\ref{fig:hyper}.
The parameter $\lambda$ controls the extent of frequency mixing between the source and target domains, while the thresholds $\tau_l$ and $\tau_h$ ensure cross-domain consistency alignment while preserving domain-specific variations.
Our study finds that a moderate frequency mixing ratio and properly selected alignment thresholds ensure effective cross-domain time series imputation.
\begin{figure*}[ht]
\centering
\includegraphics[width=0.85\linewidth]{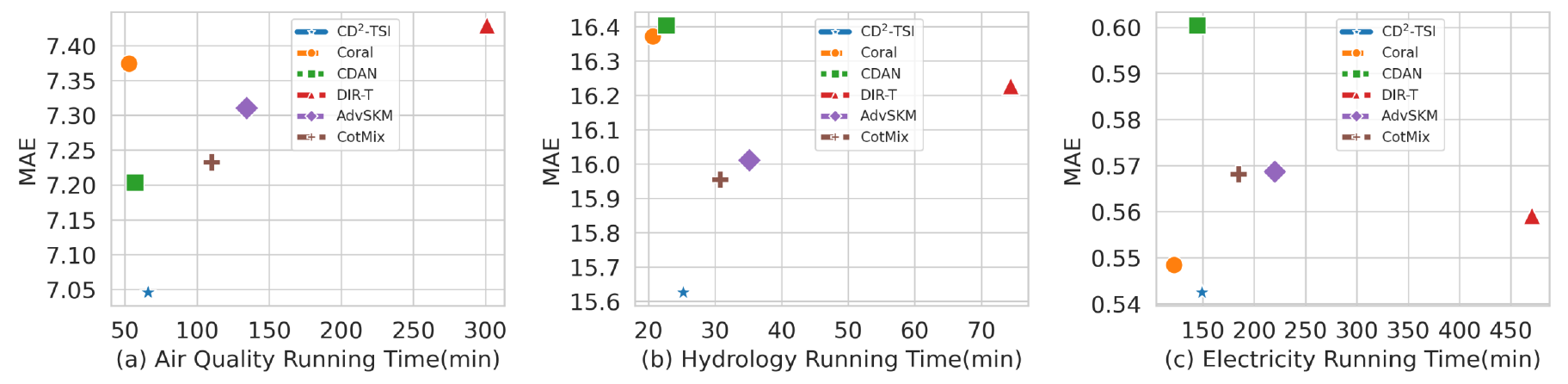} 
\caption{Efficiency analysis on Air Quality, Hydrology and Electricity datasets.}
\label{fig:efficiency}
\end{figure*}

\noindent \textbf{Efficiency Study.}
We illustrate the total training time of DA models trained on all three datasets, and the experiments are conducted on an NVIDIA RTX 4090 GPU with 24G memory. Fig.~\ref{fig:efficiency} shows the Time-MAE curve, indicating the relationship between time complexity and model performance. Compared with models such as Coral and CDAN, which achieve the least running time, CD$^2$-TSI
achieves better imputation results by leveraging frequency mixup and cross-domain adaptation at the cost of marginally increased training time.

\section{Conclusion}
In this paper, we introduce CD$^2$-TSI, a novel approach for cross-domain time series imputation, addressing the limitations of existing methods in handling high missing rates and domain shifts. 
Our approach effectively leverages cross-domain information through a diffusion-based framework while preserving domain-specific temporal dependencies.
The proposed frequency mixup interpolation and selective consistency alignment strategies contribute to improved adaptation and imputation accuracy. 
CD$^2$-TSI has demonstrated superior performance on three real-world datasets through comprehensive experiments. 
Future work will explore more challenging real-world conditions with extreme missing rates and complex domain shifts.
\label{sec:conclusion}

%\newpage 
\begin{credits}
\subsubsection{\ackname} This work was supported in part by USACE under Grant No. GR40695, "Designing nature to enhance resilience of built infrastructure in western US landscapes", and by the National Science Foundation under Grant No. 2311716, "CausalBench: A Cyberinfrastructure for Causal-Learning Benchmarking for Efficacy, Reproducibility, and Scientific Collaboration".

\subsubsection{\discintname}
The authors have no competing interests to declare that are relevant to the content of this article.
\end{credits}

\bibliography{reference}
\bibliographystyle{splncs04}

%% Note that this preceding line implies that you store your BibTeX references in a file called 'mybibliography.bib'. If you instead store your references in a file with a different name, for instance 'references.bib', the preceding line should read '\bibliography{references}'. Whatever you do, DO NOT put the file name extension .bib inside the \bibliography command; this will trip up LaTeX compilers. 
%
% If you do not want to use BibTeX, you can also type up the bibliography exactly as you see fit, using the following structure:

\end{document}